%% file: main.tex
\documentclass[runningheads]{llncs}

\usepackage{eccv}

\usepackage{eccvabbrv}

\usepackage{graphicx}
\usepackage{booktabs}

\usepackage[accsupp]{axessibility}  

\usepackage[pagebackref]{hyperref}

\usepackage{orcidlink}

\usepackage{nicefrac} 
\usepackage{tikz}
\usepackage{amsmath,amssymb}
\usetikzlibrary{arrows.meta, positioning, calc, fit,
                 shadows.blur, decorations.pathreplacing}
\usepackage{multirow} 

\usepackage{xparse}
\newcommand{\TargetTag}{\mathrm{tgt}}
\newcommand{\ReferenceTag}{\mathrm{ref}}
\NewDocumentCommand{\annotvar}{m m o}{
  \IfNoValueTF{#3}
    {#1^{#2}}                
    {#1_{#3}^{#2}}           
}
\NewDocumentCommand{\Target}{m o}{
  \annotvar{#1}{\TargetTag}[#2]
}
\NewDocumentCommand{\Reference}{m o}{
  \annotvar{#1}{\ReferenceTag}[#2]
}

\usepackage{colortbl}

\begin{document}

\title{Few-Shot Personalized Age Estimation} 


\author{Jakub Paplhám\inst{1}\orcidlink{0009-0009-7110-293X} \and
Vojtěch Franc\inst{1}\orcidlink{0000-0001-7189-1224} \and
Artem Moroz\inst{2}\orcidlink{0009-0007-5831-7106}}

\authorrunning{J.~Paplhám et al.}

\institute{Department of Cybernetics, Faculty of Electrical Engineering, CTU in Prague \and Czech Institute of Informatics, Robotics and Cybernetics, CTU in Prague\\ \email{\{paplhjak, xfrancv\}@fel.cvut.cz,\ morozart@cvut.cz}}

\maketitle

\input{sections/abstract}
\input{sections/introduction}
\input{sections/related}
\input{sections/benchmark}
\input{sections/baselines}
\input{sections/experiments}
\input{sections/conclusion}

\clearpage
%
%
\bibliographystyle{splncs04}
\bibliography{main}
\end{document}

%% file: sections/abstract.tex
\begin{abstract}
Existing age estimation methods treat each face as an independent sample, learning a global mapping from appearance to age. This ignores a well-documented phenomenon: individuals age at different rates due to genetics, lifestyle, and health, making the mapping from face to age identity-dependent. When reference images of the same person with known ages are available, we can exploit this context to personalize the estimate. The only existing benchmark for this task (NIST FRVT) is closed-source and limited to a single reference image. In this work, we introduce \textit{OpenPAE}, the first open benchmark for $N$-shot personalized age estimation with strict evaluation protocols. We establish a hierarchy of increasingly sophisticated baselines: from arithmetic offset, through closed-form Bayesian linear regression, to a conditional attentive neural process. Our experiments show that personalization consistently improves performance, that the gains are not merely domain adaptation, and that nonlinear methods significantly outperform simpler alternatives. We release all models, code, protocols, and evaluation splits.
\keywords{Age Estimation \and Personalization \and Few-Shot Learning}
\end{abstract} 

%% file: sections/introduction.tex
\section{Introduction}
\label{sec:intro}

Face-based age estimation is a mature problem in computer vision. Standard approaches learn a global mapping from facial appearance to age, treating each face as an independent sample drawn from a shared population. While effective on average, this formulation ignores a fundamental reality of human biology: people age at different rates. The difference between a person's apparent age and their chronological age is heavily modulated by genetics, health, and lifestyle. Consequently, the mapping from appearance to age is not identity-invariant.

In certain applications, reference images of a target individual with known ages are available. In border control and forensics, enrollment photographs are stored alongside document metadata. In medical records and personal archives, photographs of the same person naturally span decades. A natural question follows: can we exploit these reference images to personalize and improve the age estimate of a new target image?

Despite the utility of this task, the community lacks a standardized way to evaluate it. The only existing benchmark for reference-conditioned age estimation is the NIST Face Analysis Technology Evaluation (FATE) \cite{Hanaoka2024NISTIR8525}. However, this benchmark uses a closed-source protocol with proprietary data and is limited to exactly one reference image ($N{=}1$). While NIST reports that certain proprietary submissions achieve significant accuracy gains when utilizing a reference, the academic community has no way to verify these results, inspect the models, or build upon them.

In the academic literature, several works have explored conditioning age predictions on reference faces, but with a crucial distinction: the references are drawn from \emph{arbitrary} identities rather than the same person. Methods such as MWR~\cite{shin2022moving} and DAR~\cite{sandhaus2025relativeageestimationusing} retrieve similar-looking anchor faces from the training population to calibrate predictions. While this provides a population-level correction, it does not achieve identity-specific personalization. The closest related work is MetaAge~\cite{li2022metaage}, which generates personalized regression weights conditioned on a facial-recognition embedding. However, MetaAge extracts this embedding directly from the \emph{target} image itself, rather than from references. This assumes that a static identity embedding deterministically encodes the dynamic aging trajectory of a person; an assumption at odds with the fact that recognition embeddings are explicitly trained to be invariant to age and lifestyle variations.

To address this gap, we introduce \textit{OpenPAE}, the first open benchmark for $N$-shot personalized age estimation. We formally define the task as follows: given a target face $\Target{x}$ and a context set of $N$ reference images of the \emph{same individual} with known ages, $D = \{(\Reference{x}[1], \Reference{y}[1]), \dots, (\Reference{x}[N], \Reference{y}[N])\}$, the goal is to predict the target age, $\Target{y}$. We treat this as a few-shot conditional prediction problem, where the reference set provides evidence of an individual's unique aging pattern.

To ensure rigorous and reproducible evaluation, \textit{OpenPAE} provides standardized, identity-disjoint data splits across multiple datasets. This allows researchers to evaluate methods consistently, explicitly separating the effects of true identity personalization from test-time domain adaptation.

To provide a strong foundation for the benchmark, we implement and evaluate several baseline methods that characterize what level of modeling is necessary for effective personalization. These range from a simple arithmetic offset that corrects identity-specific bias of a global model, to a closed-form Bayesian linear regression, and finally to a deterministic attentive neural process that aggregates visual evidence from an arbitrary number of references in a single forward pass.

\paragraph{\textbf{Contributions:}}
\begin{itemize}
    \item We formalize the task of $N$-shot personalized age estimation and release \textit{OpenPAE}, the first public benchmark specifically designed for this problem.
    \item We establish a set of strong baselines---ranging from simple arithmetic offsets to deep learning models---providing a foundation for future research.
    \item We provide an extensive empirical analysis, demonstrating how performance scales with the number of references $N$, and separating the effects of true personalization from test-time domain adaptation.
    \item To foster reproducible research we publicly release all experiment code, trained models, evaluation protocols and the shared evaluation API code.
\end{itemize}

The remainder of the paper is organized as follows. \Cref{sec:related} reviews related work. \Cref{sec:benchmark} formalizes the task and describes the benchmark design. \Cref{sec:baselines} details the baseline methods. \Cref{sec:experiments} reports our experimental results and analysis, and \Cref{sec:conclusion} concludes with a discussion.

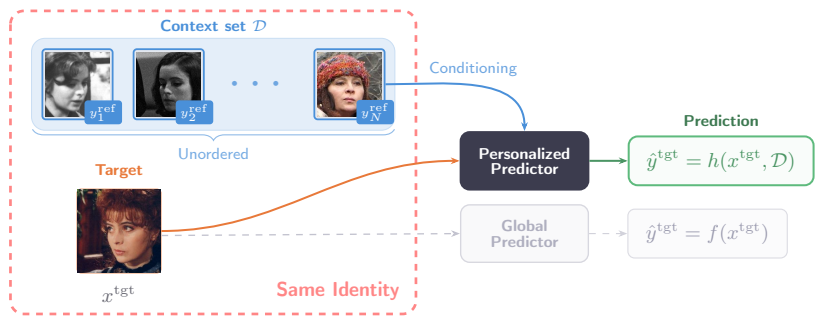
\begin{figure}
    \centering
    \resizebox{0.9\textwidth}{!}{
        \input{figures/predictor_vis}
    }
    \caption{\textbf{Task overview.}\, A global predictor (bottom) estimates age from a single face without considering individual aging rates. In personalized age estimation (top), the predictor is additionally conditioned on an unordered context set $\mathcal{D}$ of reference images with known ages from the same identity, allowing it to adapt to the individual.}
\end{figure}

%% file: figures/predictor_vis.tex
\definecolor{refblue}{HTML}{4A90D9}
\definecolor{targetorange}{HTML}{E8793A}
\definecolor{predictorgray}{HTML}{3C3C4E}
\definecolor{outputgreen}{HTML}{5EBA7D}
\definecolor{globalgray}{HTML}{9B9BAF}
\definecolor{imgfill}{HTML}{E4E8EF}
\definecolor{textdark}{HTML}{2C2C3A}
\definecolor{identitygray}{HTML}{8890A0}

\begin{tikzpicture}[
    >=Stealth,
    font=\sffamily\small,
    imgbox/.style={
        draw=#1, line width=0.8pt, rounded corners=1.5pt,
        fill=imgfill, minimum width=1.1cm, minimum height=1.1cm,
        inner sep=0pt,
    },
    agelabel/.style={
        font=\sffamily\tiny\bfseries, text=white,
        fill=refblue, rounded corners=1.5pt, 
        inner xsep=2pt, inner ysep=1.5pt,
    },
    myarrow/.style={
        -{Stealth[length=4pt, width=3pt]}, line width=0.8pt, color=#1,
    },
    trim left=-0.2cm,
]


\fill[refblue!15, draw=refblue!30, line width=0.4pt, rounded corners=6pt]
    (-0.72, 0.72) rectangle (4.97, -0.72);

\draw[pink!70!red, line width=1.2pt, dashed, rounded corners=6pt]
    (-1.05, 1.15) rectangle (5.3, -3.6);
\node[font=\sffamily\small\bfseries, text=pink!70!red, anchor=south east,
      xshift=-4pt, yshift=3pt]
    at (5.3, -3.6) {Same Identity};

\node[imgbox=refblue] (ref1) at (0, 0) {\includegraphics[width=1cm]{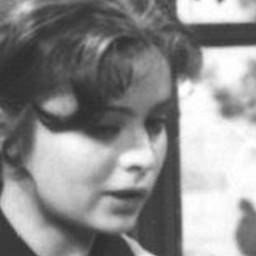}};
\node[imgbox=refblue, right=0.3cm of ref1] (ref2) {\includegraphics[width=1cm]{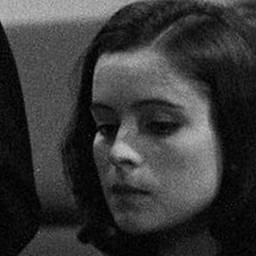}};
\node[font=\huge, text=refblue!80, right=0.25cm of ref2] (dots) {$\cdots$};
\node[imgbox=refblue, right=0.25cm of dots] (refk) {\includegraphics[width=1cm]{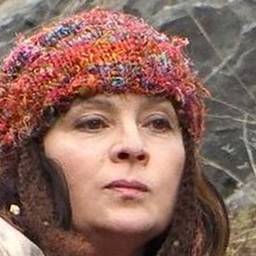}};

\node[agelabel, anchor=south east] at ([shift={(4pt,-2pt)}]ref1.south east) {$\Reference{y}[1]$};
\node[agelabel, anchor=south east] at ([shift={(4pt,-2pt)}]ref2.south east) {$\Reference{y}[2]$};
\node[agelabel, anchor=south east] at ([shift={(4pt,-2pt)}]refk.south east) {$\Reference{y}[N]$};

\coordinate (refcenter) at ($(ref1.north)!0.5!(refk.north)$);
\node[font=\sffamily\scriptsize\bfseries, text=refblue] 
    (contextlabel) at ([yshift=10pt]refcenter) {Context set $\mathcal{D}$};

\draw[decorate, decoration={brace, amplitude=5pt, mirror}, 
      refblue!60, line width=0.5pt]
    ([shift={(-4pt,-4pt)}]ref1.south west) -- ([shift={(4pt,-4pt)}]refk.south east)
    node[midway, below=5pt, font=\sffamily\scriptsize, text=refblue!75] {Unordered};

\node[imgbox=targetorange] (target) at (0.65, -2.3) {\includegraphics[width=1.3cm]{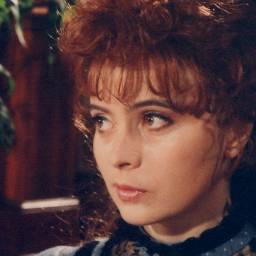}};

\node[font=\sffamily\scriptsize\bfseries, text=targetorange] 
    (targetlabel) at ([yshift=8pt]target.north) {Target};

\node[font=\sffamily\small\bfseries, text=textdark!70, below=2pt of target] (targetmath) {$\Target{x}$};

\node[
    draw=predictorgray, line width=0.9pt, rounded corners=4pt,
    fill=predictorgray, text=white,
    minimum width=2cm, minimum height=0.9cm,
    font=\sffamily\scriptsize\bfseries,
    blur shadow={shadow blur steps=4, shadow xshift=0.3mm, shadow yshift=-0.3mm},
    text width=1.7cm, align=center,
] (predictor) at (7., -1.2) {Personalized\\[-1pt]Predictor};

\node[
    draw=globalgray!40, line width=0.6pt, rounded corners=4pt,
    fill=globalgray!6, text=globalgray!80,
    minimum width=2cm, minimum height=0.9cm,
    font=\sffamily\scriptsize\bfseries,
    text width=1.7cm, align=center,
] (global) at (7., -2.34) {Global\\[-1pt]Predictor};

\node[
    draw=outputgreen, line width=0.9pt, rounded corners=4pt,
    fill=outputgreen!4,
    inner xsep=8pt, inner ysep=6pt,
    right=0.6cm of predictor,
] (output) {\textcolor{outputgreen!80!black}{$\Target{\hat{y}} = h(\Target{x}, \mathcal{D})$}};

\node[font=\sffamily\scriptsize\bfseries, text=outputgreen!65!black, anchor=south] 
    at ([yshift=1pt]output.north) {Prediction};

\node[
    draw=globalgray!35, line width=0.5pt, rounded corners=3pt,
    fill=globalgray!4,
    inner xsep=8pt, inner ysep=6pt,
    right=0.6cm of global,
] (globalout) {\textcolor{globalgray}{$\Target{\hat{y}} = f(\Target{x})$}};


\draw[myarrow=refblue, line width=0.9pt]
    (refk.east)
    to[out=0, in=90]
    node[pos=0.25, above, font=\sffamily\scriptsize, text=refblue!80]
    {\quad\qquad\qquad Conditioning}
    (predictor.north);

\draw[myarrow=targetorange, line width=0.9pt]
    (target.east) to[out=0, in=180] (predictor.west);

\draw[myarrow=globalgray!60, line width=0.6pt, dashed] 
    ([yshift=-2pt]target.east) -- (global.west);
    
\draw[myarrow=outputgreen!80!black, line width=0.9pt] 
    (predictor.east) -- (output.west);

\draw[myarrow=globalgray!45, line width=0.5pt, dashed] 
    (global.east) -- (globalout.west);

\end{tikzpicture}

%% file: sections/related.tex
\section{Related Work}
\label{sec:related}
\subsection{Age Estimation}
Age estimation from facial images has been studied extensively, with contemporary methods generally focusing on modifying the predictive head or objective function to better capture the continuous, ordinal nature of human age. Conventional classification approaches~\cite{7406390} treat discrete ages as independent classes, often computing the expected value of the softmax distribution to obtain a continuous prediction. To address the inherent ambiguity in apparent age, label distribution learning methods~\cite{7890384, gaoDLDLv2} encode the ground truth as a normal or double-exponential distribution. Ordinal regression frameworks~\cite{orcnn, coral} decompose the problem into a series of binary comparisons, while adaptive distribution methods~\cite{meanvariance} jointly optimize for prediction accuracy and calibrated variance. 

Despite their architectural and algorithmic diversity, these methods share a fundamental limitation: each face is treated as an independent sample, and the learned mapping from appearance to age is shared across the entire population. However, medical and geometric studies demonstrate that aging trajectories are heavily modulated by lifestyle, genetics, disease, and sex~\cite{bontempi2024faceage, menopause_paper}, rendering global mappings inherently limited. A notable early exception was the AGES method~\cite{ages}, which argued that age estimation should model individual aging patterns rather than treating faces as isolated samples. However, it relied on constructing explicit, dense per-identity aging sequences from hand-crafted features; an approach that scales poorly to large datasets and modern deep representations. Consequently, standard benchmarks for age estimation~\cite{morph, agedb, utkface} continue to evaluate only global, non-personalized prediction error. \textit{OpenPAE} complements these by introducing a protocol specifically designed to measure the benefit of identity-specific context.

\subsection{Reference-Conditioned Age Estimation}

Several works have explored conditioning age predictions on reference faces to improve accuracy, but with a crucial distinction from our setting: the references in the prior literature are drawn from \emph{arbitrary} identities, not the \emph{target} individual. Early approaches used comparisons against fixed global references, such as Ranking SVM~\cite{Cao2012HumanAE} or CRCNN~\cite{abousaleh2016crcnn}. More recent methods retrieve dynamic anchors from the training set: Siamese networks~\cite{jeong2018siamese} estimate age via k-nearest neighbors in a learned embedding space, MWR~\cite{shin2022moving} iteratively refines the prediction by interpolating the target's age between two references, and DAR~\cite{sandhaus2025relativeageestimationusing} predicts relative age differences against similar-looking faces. While these methods successfully calibrate predictions against global references, they do not personalize the estimate to a specific individual's aging trajectory.

The closest prior work to ours is MetaAge~\cite{li2022metaage}, which personalizes age estimation using a face recognition embedding extracted directly from the \emph{target} image. Without access to external reference images, the model effectively assumes that individuals who look alike also age alike—delegating personalization entirely to appearance similarity in the embedding space. In contrast, our approach leverages reference images of the same person at known ages, providing the model with evidence of how the individual has actually aged over time—reflecting the cumulative effects of lifestyle, health, and other factors that a single-image embedding cannot capture.

While the academic literature currently lacks methods that properly condition on same-identity references, this paradigm is already recognized in the industry. The NIST Face Analysis Technology Evaluation (FATE)~\cite{Hanaoka2024NISTIR8525} provides the only existing protocol for \textit{reference-conditioned} age estimation. However, it relies on proprietary data, evaluates submissions in a closed-source setting, and restricts the context to exactly one reference image. Consequently, the research community is left without a reproducible way to evaluate and improve upon this class of models. \textit{OpenPAE} addresses this gap by providing the first open, multi-reference benchmark for the task.

%% file: sections/benchmark.tex
\section{The \textit{OpenPAE} Benchmark}
\label{sec:benchmark}

To facilitate research in personalized age estimation, we introduce the \textit{OpenPAE} benchmark. This benchmark provides standardized evaluation protocols, multi-reference test samples, and identity-balanced metrics across four diverse datasets.

\subsection{Task Definition}
\label{ssec:task}

Given a target face image $\Target{x}\in\mathcal{X}$ of identity $i$ and a context set $\mathcal{D} = \{(\Reference{x}[j], \Reference{y}[j])\}_{j=1}^{N}$ consisting of $N$ reference images of the same identity with known chronological ages, the goal is to predict the chronological age $\Target{y}\in\mathcal{Y}\subset \mathbb{R}$ of the target. Formally, the task requires learning a predictor $h$ such that:
\begin{equation}
    \Target{\hat{y}} = h(\Target{x}, \mathcal{D}).
\end{equation}
The context set size $N$ ranges from $N=0$; reducing the task to standard global age estimation, $\Target{\hat{y}} = f(\Target{x})$; up to a dataset-dependent maximum $N=N_{\text{max}}$. No assumption is made about the temporal ordering of the images; reference images may depict the individual at ages \emph{older or younger} than the target.

\subsection{Datasets}
\label{ssec:datasets}
The benchmark evaluates performance across four established face datasets, selected for having identity annotations, sufficiently precise age labels, and multiple images per identity. \Cref{tab:datasets} summarizes the key statistics.

\begin{itemize}
    \item \textbf{CSFD-1.6M~\cite{csfd}} is the primary dataset, originally released for photo dating via facial age aggregation and repurposed here for personalized age estimation. Derived from movie and television stills, the dataset contains annotated face crops spanning decades of cinema. To ensure strict cross-dataset evaluation, we aggressively filtered CSFD to remove any identities present in other datasets, entirely removing 431 identities (64,146 images) overlapping with AgeDB. The remaining 45,989 identities were partitioned into an {\em identity-disjoint} split: 41,060 identities serve as the sole training set for all personalization methods, while 4,929 identities form the in-domain test set.
    
    \item \textbf{MORPH~\cite{morph}} is a widely used benchmark in age estimation, consisting of mugshot-style photographs. As shown in \Cref{tab:datasets}, it exhibits severe identity imbalance, making identity-balanced metrics essential. It also introduces a severe distribution shift compared to other age estimation datasets.
    
    \item \textbf{AgeDB~\cite{agedb}} is an in-the-wild dataset collected from the web, featuring celebrities with unconstrained pose, illumination, and image quality. We explicitly ensure that identities found in AgeDB are removed from CSFD in order to strictly evaluate cross-dataset generalization.
    
    \item \textbf{KANFace~\cite{kanface}} is a manually annotated dataset introduced for investigating demographic bias, captured under diverse in-the-wild conditions.
\end{itemize}

\begin{table}
\centering
\caption{Benchmark statistics. CSFD serves as the in-domain test set (Regime~B); the remaining datasets evaluate cross-dataset generalization (Regime~A).}
\label{tab:datasets}
\small
\begin{tabular}{@{}l@{\hspace{0.3cm}}r@{\hspace{0.3cm}}rrr@{\hspace{0.3cm}}r@{\hspace{0.3cm}}c@{}}
& & \multicolumn{3}{c}{\textbf{Images / ID}} & & \\
\cmidrule(lr){3-5}
\textbf{Dataset} & \textbf{Train IDs} & \textbf{Med.} & \textbf{Mean} & \textbf{Max} & \textbf{Train Imgs} & \textbf{Age Range} \\
\midrule
CSFD    & 41,060  & 7 & 33 & 1925 & 1,189,660 & [0, 110] \\
\midrule
& \textbf{Test IDs} & & & & \textbf{Test Tasks} & \\
\cmidrule(lr){2-2} \cmidrule(lr){6-6}
CSFD    & 4,929  & 11 & 40 & 1286 & 200,065           & [0, 102] \\
MORPH   & 13,158 &  3 &  4 &  52  & 3$\times$54,628   & [16, 77] \\
AgeDB   &    566 & 24 & 29 & 138  & 3$\times$16,481   & [1, 101] \\
KANFace &  1,043 & 28 & 39 & 261  & 3$\times$41,016   & [0, 100] \\
\bottomrule
\end{tabular}
\end{table}

\subsection{Evaluation Protocols}
\label{ssec:protocols}

We define two evaluation regimes to separate sources of performance gain.
\begin{itemize}
\item \textbf{Regime~A (Cross-Dataset):} Personalization models are trained exclusively on CSFD and evaluated on entire MORPH, AgeDB, and KANFace datasets without any target-domain adaptation. In this regime, the personalization mechanism may simultaneously perform domain adaptation (\eg, movie stills to mugshots) and adapt to the specific individual.
\item \textbf{Regime~B (In-Domain):} Models are evaluated on the official \textit{OpenPAE} CSFD test split, which is strictly identity-disjoint from the CSFD training set. Because the training and test data share the exact same domain distribution, any accuracy improvement over the global baseline can be attributed to identity-specific personalization rather than domain adaptation.
\end{itemize}

\paragraph{Cross-Source References} To prevent models from exploiting scene-level information on CSFD (\eg, the lighting, film grain, makeup or clothes used in a specific movie scene), Regime~B enforces a cross-source rule. Whenever possible (98\%), the reference images for a given target are drawn from a \emph{different} movie.

\paragraph{Fixed Evaluation Tasks} Evaluating all possible reference combinations for identities with many images leads to a combinatorial explosion (\eg, over $10^{10}$ ways to select 10 references from 50 images). To circumvent this and ensure exact reproducibility, \textit{OpenPAE} provides all evaluation tasks as a fixed list. Each test task is formalized as a tuple $(\Target{x}, \mathcal{D}_{\text{max}}, \Target{y})$, where $\Target{x}$ is the target image, $\Target{y}$ is the ground truth age, and $\mathcal{D}_{\text{max}}$ is a pre-sampled, randomly ordered list of available reference images and their corresponding ages for that identity. Evaluation at any specific reference count $N$ is achieved by strictly truncating $\mathcal{D}_{\text{max}}$ to its first $N$ entries. Due to the large scale of CSFD, a single randomized trial over the dataset yields statistically stable estimates. For the smaller datasets (AgeDB, MORPH, KANFace), we use three distinct $\mathcal{D}_{\text{max}}$ orderings per target $\Target{x}$ and report the averaged results.

\subsection{Metrics}
\label{ssec:metrics}

Standard mean absolute error (MAE) weights each image equally. In long-tailed datasets, this allows frequently photographed individuals to skew the evaluation. To address this, our primary metric is identity-averaged MAE:
\begin{equation}
    \text{MAE}_{\text{id}} = \frac{1}{|\mathcal{I}|} \sum_{i \in \mathcal{I}} \left( \frac{1}{|\mathcal{T}_i|} \sum_{t \in \mathcal{T}_i} \left| \Target{y} - \Target{\hat{y}} \right| \right)
\end{equation}
where $\mathcal{I}$ is the set of all identities and $\mathcal{T}_i$ is the set of test target images for identity $i\in\mathcal{I}$. This metric first computes the mean absolute error within each identity and then averages across all identities, weighting every individual equally regardless of their representation in the dataset. 

%% file: sections/baselines.tex
\section{Baseline Methods}
\label{sec:baselines}

To establish a comprehensive foundation for \textit{OpenPAE}, we design a suite of baselines that isolate different mechanisms of personalization. These range from a standard global estimator, to scalar bias correction, closed-form statistical calibration, and deep sequence models.

\subsection{Global Age Estimator (Global)}
\label{ssec:baseline_global}

All personalized methods are grounded in a non-personalized, global age estimator. We employ a ViT-B/16 backbone initialized with FaRL~\cite{farl}, pretrained via visual-linguistic supervision on facial images. The network extracts an embedding ($d{=}512$) from the CLS token, which is passed to a linear head that outputs a predicted mean age $\mu$ and a log-variance $\log \sigma^2$. The entire model is finetuned on the CSFD dataset by minimizing the Gaussian negative log-likelihood: $\mathcal{L} = \nicefrac{(y - \mu)^2}{2\sigma^2} + \nicefrac{1}{2}\log\sigma^2$. The \textbf{Global} model represents the standard age estimation paradigm: given a target face $\Target{x}$, it predicts $\Target{\hat{y}} = f(\Target{x})$ without access to any references.

\subsection{Arithmetic Calibration (Offset)}
\label{ssec:baseline_offset}

The simplest form of personalization assumes that the global model makes a systematic error across images of the same individual. If the model consistently overestimates a person's age on the reference images, we can apply an opposite correction to the target. Formally, \textbf{Offset} computes the mean residual of the \textit{Global} model on the reference set $\mathcal{D}$ and applies it as an additive correction: $\Target{\hat{y}} = f(\Target{x}) + \frac{1}{N} \sum_{j=1}^{N} \big(\Reference{y}[j] - f(\Reference{x}[j])\big)$. While this requires no additional training, the assumption of a constant chronological bias is fragile. 


\subsection{A Probabilistic Framework for Personalization}
\label{ssec:baseline_generative}

To move beyond simple scalar offsets, we formalize personalized age estimation through a hierarchical Bayesian generative model. We posit that each identity ${i\in\mathcal{I}}$ is associated with a latent parameter $\theta_i \sim p(\theta)$ drawn from a population-level prior. This parameter encapsulates the unobserved biological and lifestyle factors that govern the identity-specific mapping from facial appearance to perceived age. Given an image $x$ of identity $i$, the age label is drawn as ${y \sim p(y \mid x, \theta_i)}$.

At test time, we are given a target image $\Target{x}$ and a reference set $\mathcal{D} = \{(\Reference{x}[j], \Reference{y}[j])\}_{j=1}^{N}$. Our goal is to compute the posterior predictive distribution:
\begin{equation}
  p(\Target{y} \mid \Target{x}, \mathcal{D})
  = \int p(\Target{y} \mid \Target{x}, \theta)\,
        p(\theta \mid \mathcal{D})\, d\theta
  \label{eq:predictive}
\end{equation}
The remaining baselines implement this framework in two distinct ways: Bayesian Linear Regression (\textbf{BLR}, \Cref{ssec:baseline_blr}) computes this integral in closed form under strong assumptions, while the deep sequence models (\Cref{ssec:baseline_deep}) avoid the integral by including $\mathcal{D}$ in the forward pass and directly approximating the predictive distribution.

\subsection{Bayesian Linear Regression (BLR) \cite{blr}}
\label{ssec:baseline_blr}
Let $\phi(x) \in \mathbb{R}^{d+1}$ denote the one-augmented feature vector extracted by the frozen Global backbone. The Global baseline implicitly defines a weight vector $\theta_{\text{global}}$ such that $f(x) = \theta_{\text{global}}^\top \phi(x)$. To personalize the prediction, we associate each identity $i$ with a distinct linear head $\theta_i$. We place a diagonal Gaussian prior on this identity-specific head, centered on the global weights: 
\begin{equation}
\theta_i \sim \mathcal{N}(\theta_{\text{global}}, \mathrm{diag}(\mathbf{s}^2))
\end{equation}
where $\mathbf{s}^2 \in \mathbb{R}^{d+1}$. Rather than assuming homoscedatic aleatoric noise, we leverage the Global network's uncertainty estimate. The noise variance for image $x$ is modeled as $\sigma^2(x) = \gamma_1 \cdot \hat{v}(x) + \gamma_2$, where $\hat{v}(x)$ is the variance output by the Global head. This preserves the learned uncertainty $\hat{v}(x)$ while correcting for systematic miscalibration. The prior hyperparameters $(\mathbf{s}^2, \gamma_1, \gamma_2)$ are estimated via empirical Bayes, maximizing marginal log-likelihood across training identities. 

At test time, both the posterior $p(\theta_i \mid \mathcal{D})$ and the predictive distribution $p(\Target{y} \mid \phi(\Target{x}), \mathcal{D})$ can be computed in closed form. The final point prediction $\Target{\hat{y}}$ is taken as the mean of this predictive distribution. With no reference data, the posterior mean falls back to the global predictor $\theta_{\text{global}}$. As $N$ grows, the references pull the posterior toward the identity-specific least-squares solution.

\subsection{Deep Sequence Models}
\label{ssec:baseline_deep}

While BLR provides a rigorous closed-form solution, it assumes a linear relationship between deep features and identity-specific aging. To capture the highly nonlinear visual transformations associated with human aging, deep sequence models offer a flexible alternative. Processing a reference set requires permutation invariance, as a collection of $N$ reference images possesses no inherent sequential order. Attention-based architectures naturally satisfy this requirement while accommodating variable-sized context sets in the forward pass. We evaluate two distinct attention-based baselines.

\subsubsection{Attention Global (Attn-G)} 
tests the benefit of nonlinearity in strict isolation. It operates on the exact same CLS-token embedding as the Global model and BLR. Because the representation is compact (one token per image), the context set can be processed using unrestricted self-attention. All reference tokens, each summed with a sinusoidal age embedding, are concatenated with the target token into a single sequence and processed by a Transformer encoder. A crucial masking strategy is applied: reference tokens can attend to each other to build context, but cannot attend to the target, while the target token attends to the full sequence. The target output token is then decoded into $\Target{\hat{y}} = \mu$ and $\log\sigma^2$ via linear heads, trained with the same NLL objective as the Global model. 

\subsubsection{Attention Spatial (Attn-S)} 
extends the approach to operate on spatial features. It extracts patch features from intermediate ViT layers, yielding a grid of $14\times 14 = 196$ spatial tokens per image. Applying the unrestricted masked self-attention of Attn-G to this expanded representation is computationally prohibitive, scaling quadratically as $\mathcal{O}((N \cdot 196)^2)$. Therefore, Attn-S adopts a cross-attention bottleneck. The 196 tokens of each reference image, summed with their respective sinusoidal age embeddings, are flattened into a unified memory bank of length $N \times 196$. The target spatial tokens act as Queries in a Multi-Head Cross-Attention layer, retrieving age-conditioned features from the reference memory bank. The retrieved features are refined with self-attention, globally pooled, and passed to the prediction heads as in Attn-G, again setting $\Target{\hat{y}} = \mu$.

\paragraph{Theoretical Interpretation}
Both architectures can be interpreted probabilistically as approximating the integral in \Cref{eq:predictive}. The causal-style masking strategy utilized in Attn-G aligns with the structural conditions identified by M\"{u}ller~\etal~\cite{muller2022pfn} for valid Bayesian estimation via Prior-Fitted Networks (PFNs). Similarly, the cross-attention bottleneck in Attn-S follows the established framework of Attentive Conditional Neural Processes~\cite{kim2019anp, garnelo2018cnp}. By incorporating $\mathcal{D}$ directly into the forward pass, these attention models can directly approximate the posterior predictive distribution without requiring closed-form solutions.

\subsection{Pairwise Averaging (Pair-Avg)}
\label{ssec:baseline_pairavg}

To isolate the value of attention over the full context, \textbf{Pair-Avg} utilizes the same trained weights as Attn-S but alters the evaluation protocol. Instead of providing the full set of $N$ references to the module simultaneously, each reference is processed individually. An independent forward pass is executed for each reference, and the arithmetic mean of the resulting predictions is computed.


%% file: sections/experiments.tex
\section{Experiments}
\label{sec:experiments}
\subsection{Implementation Details}
\label{ssec:implementation}
The faces are detected and aligned using facial landmarks. We extract a wide crop around the aligned face; preserving context such as hair, ears, and the jawline; before resizing to $256 \times 256$ pixels. During training, images are center-cropped to $224 \times 224$ and subjected to data augmentation, including random horizontal flipping ($p=0.5$), color jitter, and random grayscale conversion ($p=0.05$). Across all experiments, models are optimized using AdamW, maintaining an Exponential Moving Average (EMA) of the weights with a decay factor of $0.999$.

\begin{itemize}
    \item \textbf{Global Age Estimator:} The global ViT-B/16 baseline is trained for up to 50 epochs with a batch size of 256 images. We employ a learning rate of $10^{-5}$ for the backbone to preserve the pretrained FaRL representation, and $10^{-3}$ for the randomly initialized regression head. The model is optimized using Gaussian NLL with a weight decay of $10^{-4}$. We use 90\% of the identities for training and 10\% for validation. The trained model serves as the initialization for BLR and the Deep Sequence Models.

    \item \textbf{BLR Prior Estimation:} The empirical Bayes optimization for the BLR hyperparameters ($\mathbf{s}^2, \gamma_1, \gamma_2$) maximizes the marginal log-likelihood using full-validation-dataset gradient descent over $2,000$ iterations with Adam optimizer with a learning rate of $10^{-3}$ decayed following cosine annealing. Prior and noise variances are parameterized and optimized in logarithmic space.

    \item \textbf{Deep Sequence Models (Attn-G \& Attn-S):} Unlike the global model, the sequence models are trained using a task-based batching strategy. Each batch samples 10 distinct identities; for each identity, one image is selected as the target alongside $N \in [1, 20]$ dynamically sampled reference images. We train until up to 300,000 batches have been processed.
    
    \begin{itemize}
        \item \textit{Regularization:} We add Gaussian noise ($\sigma = 2$ years) to $\Reference{y}[j]$ of the reference set $\mathcal{D}$ during training. Gradients are clipped to unit norm.
        \item \textit{Optimization:} Same as the global baseline, these models are optimized using Gaussian NLL. We apply a learning rate of $10^{-6}$ to the pretrained backbone and $10^{-4}$ to the newly initialized Transformer modules. The prediction is bounded by a Sigmoid activation, rescaled to $[0, 100]$.
        \item \textit{Attn-S Specifics:} For the spatial model, we concatenate patch features from intermediate ViT layers 5, 8, and 11 to form the spatial grid.
    \end{itemize}
\end{itemize}

\begin{figure}
  \centering
  \includegraphics[width=0.9\linewidth]{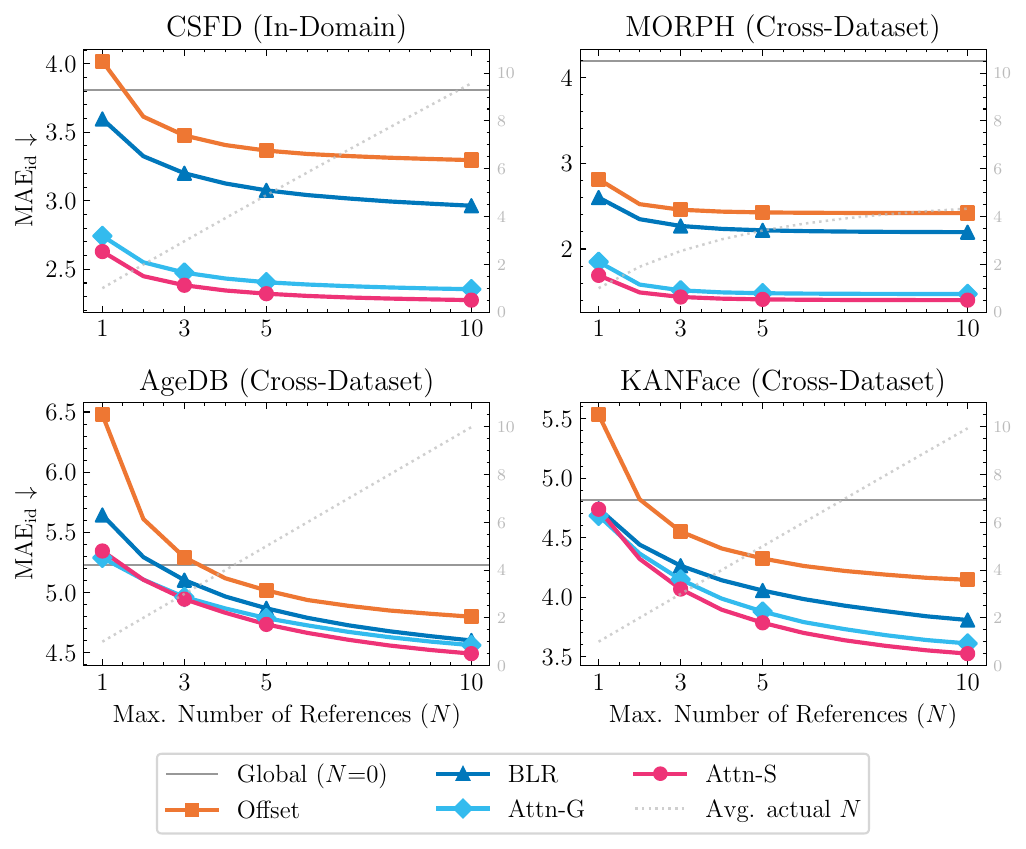}
  \caption{\textbf{OpenPAE Results.} Identity-balanced $\text{MAE}_{\text{id}} \downarrow$ as a function of the maximum number of allowed references $N$. The horizontal line marks the global baseline ($N{=}0$). The dotted line shows the average number of references actually available per identity.}
  \label{fig:n_vs_mae}
\end{figure}

\begin{figure}
  \centering
  \includegraphics[width=0.95\linewidth]{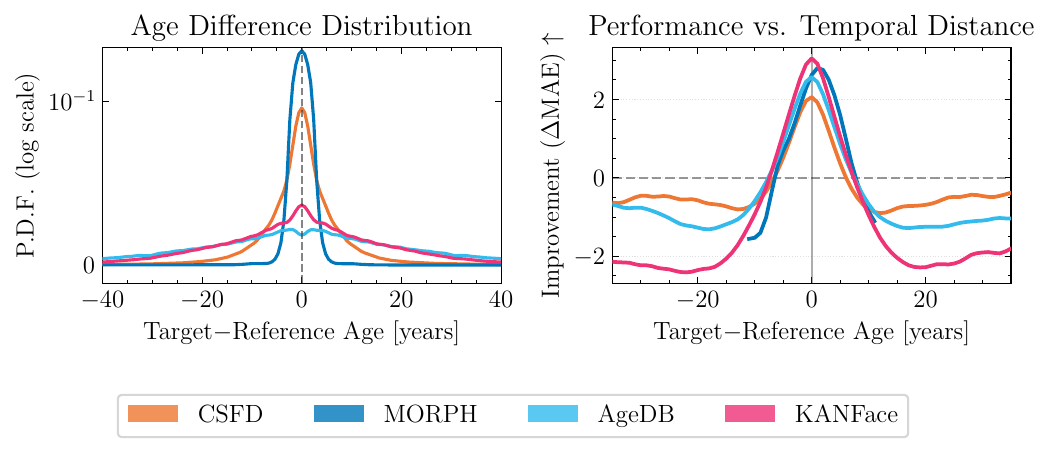}
  \caption{\textbf{Impact of Temporal Distance on Personalization.} \textbf{Left:} Probability density of the age gap between target and reference images. \textbf{Right:} Relative improvement ($\Delta \text{MAE}_{\text{id}}$) of the Pair-Avg $N=1$ model over the unpersonalized global baseline. With the age difference $|\Target{y} - \Reference{y}|$ below 10 years, the model consistently improves upon the global baseline on all datasets. When confronted with the heavy temporal tails of AgeDB or KANFace, these large age differences fall out-of-distribution, causing the personalized model to perform worse than the unconditioned global predictor.}
  \label{fig:age_diff}
\end{figure}

\input{tables/table_1}
\input{tables/table_2}

\subsection{Main Results}
\Cref{tab:main_results} and \cref{fig:n_vs_mae} present the identity-balanced $\text{MAE}_{\text{id}}$ across all datasets and context sizes $N$. We observe several key findings regarding the mechanisms necessary for effective personalized age estimation.

\textbf{Deep Feature Aggregation.} Across all datasets, the deep sequence models (Attn-G, Attn-S) outperform both the scalar bias correction (Offset) and the closed-form statistical calibration (BLR). \Eg, on the CSFD at $N=10$, Offset only reduces the $\text{MAE}_{\text{id}}$ from 3.81 to 3.29, whereas Attn-S reduces it to 2.27.

\textbf{Joint Context vs.\ Averaging.} As shown in \Cref{tab:main_results}, Pair-Avg quickly saturates; on MORPH, its performance flatlines at 1.64 $\text{MAE}_{\text{id}}$ from $N=3$ onwards. In contrast, Attn-S continues to scale down to 1.40 $\text{MAE}_{\text{id}}$. This demonstrates that using the full context set can use the informative references while suppressing noisy or uninformative ones.

\textbf{Robustness.} While conditioning on references generally improves performance, \Cref{tab:main_results} shows fragility in the single-reference ($N=1$) setting. On AgeDB, $N=1$ degrades performance for all methods compared to the global model (\eg, Attn-S from 5.23 to 5.34). However, as context size grows, the personalization consistently outperforms the global model. This underscores the importance of the proposed $N$-shot benchmark formulation compared to NIST \cite{Hanaoka2024NISTIR8525}.

\subsection{Separating Personalization from Domain Adaptation}
\label{ssec:identity_swap}
A critical question is whether any observed performance gains stem from true identity-specific modeling, or merely from test-time domain adaptation. Because reference images are drawn from the same dataset as the target, they inherently provide the model with domain-level information. To isolate the effect of identity, we design a controlled evaluation protocol that preserves the \emph{domain} and \emph{age-prior} information while entirely destroying the \textit{correct identity signal}.

For each evaluation task, we construct two alternative context sets by replacing the original same-identity references with images of different individuals. These substitute images are strictly constrained to have the exact same chronological ages as the original references.

\begin{itemize}
\item \textbf{Different ID (mix):} Each reference slot is filled by a randomly selected image of the correct age, belonging to a different identity. This produces age-matched, domain-matched context set without a consistent identity signal.
\item \textbf{Different ID (1):} All reference slots are filled by a single alternative identity whose available photos match the required chronological ages. This produces an age-matched, domain-matched context set that holds a consistent identity signal, however, of an \emph{incorrect identity}.
\end{itemize}

All swapped reference images are drawn exclusively from the test set. We apply a penalty during the search to prevent the evaluation from over-relying on a small subset of highly photographed individuals as source of the references.

This protocol establishes a strict interpretation framework for our results. If providing a wrong-identity context set improves performance over the Global baseline, that gain is attributable purely to domain and prior adaptation. Conversely, any performance gap between the correct-identity context and the two wrong-identity contexts quantifies the exact benefit of true personalization. Note that for in-domain evaluation (CSFD), the global model is domain-adapted, hence the wrong-identity contexts only inform about the identity age prior.

\subsubsection{Age Prior vs.\ True Personalization.} \Cref{tab:domain_adaptation} reports the results of this ablation. Providing references of incorrect identities but matched ages still yields substantial gains over the Global model. For example, on MORPH at $N=10$, \textit{Different ID (mix)} drops the $\text{MAE}_{\text{id}}$ from 4.19 to 1.81. We hypothesize this is because the reference ages are often close to the target age. Across all datasets at $N=10$, conditioning on the correct identity (\textit{Same ID}) yields a highly consistent additional reduction of $\sim$0.4 to 0.7 $\text{MAE}_{\text{id}}$ compared to \textit{Different ID (mix)} (\eg, 5.20 vs.\ 4.49 on AgeDB, 2.75 vs.\ 2.27 on CSFD). This is the performance gain that we attribute purely to the identity-specific personalization.

\subsubsection{Consistent Wrong Identity.} \Cref{tab:domain_adaptation} shows that conditioning on a single incorrect identity \textit{Different ID (1)} consistently yields higher error than a mixture of incorrect identities \textit{Different ID (mix)}. This validates that the model successfully adapts to the specific individual rather than relying solely on the domain or age prior. A single alternative identity introduces a consistent biological bias, which the model learns and incorrectly applies to the target. In the mixed setting the biases cancel out, isolating the pure domain and prior adaptation effects.

\subsection{The Impact of Temporal Distance}
Because the temporal distribution of reference images varies wildly by dataset, we investigate how the temporal gap between the reference and the target affects personalization. To isolate the helpfulness of the reference, \Cref{fig:age_diff}, right, plots the relative improvement over the global model $\text{MAE}_{\text{id,Global}} - \text{MAE}_{\text{id,Attn-S}}$ as a function of temporal distance $\Target{y}-\Reference{y}$ for the Pair-Avg model with $N=1$.

As expected, references with similar ages to the target provide a strong signal, improving predictions by up to 3 years. However, at temporal distances greater than 10 years, the pairwise improvement drops below zero, degrading the prediction. We attribute this to a distribution shift between the training and test domains. As shown in \cref{fig:age_diff} (Left), the CSFD training set consists almost entirely of age differences within $\pm 10$ years. Within this regime, the model consistently improves upon the global baseline on all datasets. When confronted with the heavy temporal tails of AgeDB or KANFace, these large age differences fall out-of-distribution, causing the personalized model to perform worse than the unconditioned global predictor. We pose this as an open challenge for the community: achieving temporal robustness under severe distribution shifts. The \textit{OpenPAE} benchmark explicitly allows evaluating this property by training exclusively on the CSFD dataset and testing across the diverse target datasets.

%% file: tables/table_1.tex
\begin{table}[t]
\centering
\caption{\textbf{OpenPAE Results.} Identity-balanced mean absolute error ($\text{MAE}_{\text{id}} \downarrow$ ) across all evaluated datasets and methods. $N$ indicates the number of reference images in the context set. The \emph{Global} column ($N=0$) is the standard unpersonalized baseline.}
\label{tab:main_results}
\setlength{\tabcolsep}{6pt}
\begin{tabular}{cl c cccc}
\toprule
& \multirow{2}{*}{\textbf{Method}} &  & \multicolumn{4}{c}{\textbf{Number of References ($N$)}} \\
 \cmidrule(lr){4-7}
& & $N=0$ & $N=1$ & $N=3$ & $N=5$ & $N=10$ \\
\midrule
\multicolumn{7}{c}{\scriptsize\textit{In-Domain Evaluation (Regime B)}} \\
\midrule
\multirow{5}{*}{\rotatebox[origin=c]{90}{\textbf{CSFD}}} 
& Offset   & \multirow{5}{*}{3.81} & 4.01 & 3.47 & 3.36 & 3.29 \\
& BLR      &  & 3.59 & 3.20 & 3.07 & 2.96 \\
& Attn-G   &  & 2.74 & 2.47 & 2.40 & 2.35 \\
& Pair-Avg &  & 2.62 & 2.47 & 2.44 & 2.42 \\
& Attn-S &  & 2.62 & 2.38 & 2.32 & 2.27 \\
\midrule
\multicolumn{7}{c}{\scriptsize\textit{Cross-Dataset Generalization (Regime A)}} \\
\midrule
\multirow{5}{*}{\rotatebox[origin=c]{90}{\textbf{MORPH}}} 
& Offset   & \multirow{5}{*}{4.19} & 2.81 & 2.45 & 2.42 & 2.41 \\
& BLR      &  & 2.59 & 2.26 & 2.21 & 2.19 \\
& Attn-G   &  & 1.84 & 1.51 & 1.48 & 1.47 \\
& Pair-Avg &  & 1.69 & 1.64 & 1.64 & 1.64 \\
& Attn-S &  & 1.69 & 1.44 & 1.41 & 1.40 \\
\midrule
\multirow{5}{*}{\rotatebox[origin=c]{90}{\textbf{AgeDB}}} 
& Offset   & \multirow{5}{*}{5.23} & 6.47 & 5.29 & 5.01 & 4.79 \\
& BLR      &  & 5.64 & 5.10 & 4.86 & 4.60 \\
& Attn-G   &  & 5.28 & 4.96 & 4.78 & 4.56 \\
& Pair-Avg &  & 5.34 & 4.89 & 4.81 & 4.74 \\
& Attn-S &  & 5.34 & 4.94 & 4.73 & 4.49 \\
\midrule
\multirow{5}{*}{\rotatebox[origin=c]{90}{\textbf{KANFace}}} 
& Offset   & \multirow{5}{*}{4.82} & 5.53 & 4.55 & 4.32 & 4.14 \\
& BLR      &  & 4.74 & 4.26 & 4.05 & 3.80 \\
& Attn-G   &  & 4.68 & 4.14 & 3.87 & 3.61 \\
& Pair-Avg &  & 4.73 & 4.19 & 4.08 & 4.01 \\
& Attn-S &  & 4.73 & 4.06 & 3.78 & 3.52 \\
\bottomrule
\end{tabular}
\end{table}

%% file: tables/table_2.tex
\begin{table}[t]
\centering
\caption{\textbf{Personalization or Domain Adaptation?} Table shows $\text{MAE}_{\text{id}} \downarrow$. 
\textbf{Same ID}: Standard evaluation. 
\textbf{Different ID (1)}: Each reference from one wrong identity, age-matched.
\textbf{Different ID (mix)}: Each reference from a different wrong identity, age-matched. The different ID results represent domain adaptation without personalization.}
\label{tab:domain_adaptation}
\footnotesize
\setlength{\tabcolsep}{1.5pt}
\renewcommand{\arraystretch}{1.03}
\newcolumntype{G}{>{\columncolor{gray!8}}c}
\begin{tabular}{@{}c l c @{\hskip 6pt} GGG @{\hskip 6pt} ccc @{\hskip 6pt} GGG@{}}
\toprule
& \textbf{Method} & & \multicolumn{3}{c}{\textbf{Different ID\ (1)}} 
      & \multicolumn{3}{c}{\textbf{Different ID\ (mix)}} 
      & \multicolumn{3}{c}{\textbf{Same ID}} \\
\cmidrule(lr){4-6} \cmidrule(lr){7-9} \cmidrule(lr){10-12}
& & $N\!=0$ 
  & \cellcolor{white} $N\!=1$ & \cellcolor{white}$N\!=5$ & \cellcolor{white} $N\!=10$
  & $N\!=1$ & $N\!=5$ & $N\!=10$
  & \cellcolor{white} $N\!=1$ & \cellcolor{white} $N\!=5$ & \cellcolor{white} $N\!=10$ \\
\midrule
\multicolumn{12}{c}{\scriptsize\textit{In-Domain Evaluation (Regime B)}} \\
\midrule
\multirow{3}{*}{\rotatebox[origin=c]{90}{\tiny\textbf{CSFD}}}
& BLR             & \multirow{3}{*}{3.81} & 4.50 & 4.19 & 4.05 & 4.51 & 3.81 & 3.65 & 3.59 & 3.07 & 2.96 \\
& Attn-G          &  & 3.27 & 3.08 & 2.96 & 3.11 & 2.80 & 2.75 & 2.74 & 2.40 & 2.35 \\
& Attn-S          &  & 3.30 & 3.12 & 3.00 & 3.13 & 2.81 & 2.75 & 2.62 & 2.32 & 2.27
\\
\midrule
\multicolumn{12}{c}{\scriptsize\textit{Cross-Dataset Generalization (Regime A)}} \\
\midrule
\multirow{3}{*}{\rotatebox[origin=c]{90}{\tiny\textbf{MORPH}}}
& BLR             & \multirow{3}{*}{4.19}  & 4.48 & 4.31 & 4.28 & 4.51 & 3.78 & 3.73 & 2.59 & 2.21 & 2.19 \\
& Attn-G          &  & 2.30 & 2.00 & 1.99 & 2.35 & 1.87 & 1.86 & 1.84 & 1.48 & 1.47 \\
& Attn-S          &  & 2.15 & 1.96 & 1.95 & 2.25 & 1.82 & 1.81 & 1.69 & 1.41 & 1.40 \\
\midrule
\multirow{3}{*}{\rotatebox[origin=c]{90}{\tiny\textbf{AgeDB}}}
& BLR             & \multirow{3}{*}{5.23} & 6.37 & 6.03 & 5.88 & 6.39 & 5.73 & 5.48 & 5.64 & 4.86 & 4.60 \\
& Attn-G          &  & 5.64 & 5.66 & 5.38 & 5.56 & 5.43 & 5.25 & 5.28 & 4.78 & 4.56 \\
& Attn-S          &  & 5.82 & 5.65 & 5.36 & 5.70 & 5.40 & 5.20 & 5.34 & 4.73 & 4.49 \\
\midrule
\multirow{3}{*}{\rotatebox[origin=c]{90}{\tiny \textbf{KANFace}}}
& BLR             & \multirow{3}{*}{4.82} & 5.41 & 5.30 & 5.06 & 5.43 & 4.87 & 4.66 & 4.74 & 4.05 & 3.80 \\
& Attn-G          &  & 4.98 & 4.86 & 4.39 & 4.97 & 4.54 & 4.32 & 4.68 & 3.87 & 3.61 \\
& Attn-S          &  & 5.13 & 4.88 & 4.28 & 5.09 & 4.43 & 4.21 & 4.73 & 3.78 & 3.52 \\
\bottomrule
\end{tabular}
\end{table}

%% file: sections/conclusion.tex
\section{Conclusion}
\label{sec:conclusion}
This work formally introduces the task of $N$-shot personalized age estimation. We proposed the \textit{OpenPAE} benchmark to standardize evaluation across diverse, cross-domain datasets. We provide an experimental setup to explicitly separate true identity-specific personalization from dataset bias and age prior correction. Our extensive baseline evaluation demonstrates that human aging trajectories are highly complex and identity-dependent; effectively modeling them requires deep models and joint-context aggregation rather than simple arithmetic offsets.
